\documentclass[runningheads]{llncs}
\usepackage{graphicx}
\usepackage{color}
\usepackage{xcolor}
\usepackage{amssymb}
\usepackage{amsmath}
\usepackage{float}
\usepackage{environ} 
\usepackage{lipsum} 
\usepackage{breqn}

\NewEnviron{SMALL}{%
    \scalebox{0.89}{$\BODY$} 
}
\NewEnviron{SMALLL}{%
    \scalebox{0.94}{$\BODY$} 
}

\begin{document}
\title{Automatic-differentiated \\ Physics-Informed Echo State Network (API-ESN)
\thanks{A. Racca is supported by the EPSRC-DTP and the Cambridge Commonwealth, European \& International Trust under a Cambridge European Scholarship. L. Magri is supported by the Royal Academy of Engineering Research Fellowship scheme and the visiting fellowship at the Technical University of Munich – Institute for Advanced Study, funded by the German Excellence Initiative and the European Union Seventh Framework Programme under grant agreement
n. 291763.}}
%
%
\author{Alberto Racca\inst{1} \and
Luca Magri\inst{1,2,3,4}}
\authorrunning{A. Racca and L. Magri}
%
\institute{Department of Engineering, University of Cambridge, UK
\\
\and
The Alan Turing Institute, London, UK 
\\
\and
Imperial College London, Aeronautics Department, London, UK 
\\
\and
(visiting) Institute for Advanced Study, Technical University of Munich, Germany
}
\maketitle              
\begin{abstract}
We propose the Automatic-differentiated Physics-Informed Echo State Network (API-ESN).
The network is constrained by the physical equations through the reservoir's exact time-derivative, which is computed by automatic differentiation.  
As compared to the original Physics-Informed Echo State Network, the accuracy of the time-derivative is increased by up to seven orders of magnitude. 
This increased accuracy is key in chaotic dynamical systems, where errors grows exponentially in time. 
The  network is showcased in the reconstruction of unmeasured (hidden) states of a chaotic system.
The API-ESN eliminates a source of error, which is present in existing physics-informed echo state networks, in the computation of the time-derivative. This opens up new possibilities for an accurate reconstruction of chaotic dynamical states.

\end{abstract}
\section{Introduction}
In fluid mechanics,  we only rarely have experimental measurements on the entire state of the system because of technological/budget constraints on the number and placement of sensors. In fact, we typically measure only a subset of the state, the \emph{observed} states, but we do not have data on the remaining variables, the \emph{hidden} states. 
In recent years, machine learning techniques have been proposed to infer hidden variables, which is also known as {\it reconstruction}. 
A fully-data driven approach to reconstruction assumes that data for the hidden states is available only for a limited time interval, which is used to train the network \cite{lu2017reservoir}. 
On the other hand, a physics-informed approach to reconstruction employs the governing equations \cite{doan2020learning,raissi2020hidden}. 
In this work, we use automatic differentiation to eliminate the source of error in the Physics-Informed Echo State Network, which  originates from approximating the time-derivative of the network \cite{doan2020learning}. Automatic differentiation records the elementary operations of the model and evaluates the derivative by applying the chain rule to the derivatives of these operations \cite{baydin2017automatic}. With automatic differentiation, we compute exactly the time-derivative, thereby extending the network's ability to reconstruct hidden states in chaotic systems.  
 In section \ref{sec:PIESN}, we present the proposed network: the Automatic-differentiated Physics-Informed Echo State Network (API-ESN). In section \ref{sec:Results}, we discuss the results. We summarize the work and present future developments in section \ref{sec:conc}. 

\section{Automatic-differentiated Physics-Informed Echo State Network (API-ESN)}

\label{sec:PIESN}

We study the nonlinear dynamical system 
\begin{equation}
  \label{dyn_sys}
   \bf{\dot{y}}=\mathbf{f}(\mathbf{y}),  
\end{equation}
where $\textbf{y}\in\mathbb{R}^{N_{y}}$ is the state of the physical system, 
$\textbf{f}$ is a nonlinear operator,
and $\dot{\;\;}$ is the time-derivative. 
We consider a case where $\textbf{y}$ consists of an observed state, $\textbf{x}\in\mathbb{R}^{N_{x}}$, and a hidden state, $\textbf{h}\in\mathbb{R}^{N_{h}}$: $\textbf{y}=[\textbf{x};\textbf{h}]$, where $[\cdot\,;\cdot]$ indicates vertical concatenation and $N_y=N_x+N_h$. We assume we have non-noisy data on $\mathbf{x}$, and its derivative, $\mathbf{\dot{x}}$, which can be computed offline. We wish to reconstruct $\textbf{h}$ given the data. The $N_t+1$ training data points for the observed states are $\textbf{x}(t_i)$ for $i=0,1,2,\ldots,N_t$, taken from a time series that ranges from $t_0=0$ to $t_{N_t}=N_t\Delta t$, where $\Delta t$ is the constant time step. We introduce the Automatic-differentiated Physics-Informed Echo State Network (API-ESN) to reconstruct $\textbf{h}$ at the same time instants, by constraining the time-derivative of the network through the governing equations. The network is based on the Physics-Informed Echo State Network (PI-ESN) \cite{doan2019physics,doan2020learning}, which, in turn, is based on the fully data-driven ESN \cite{jaeger2004harnessing,lukovsevivcius2012practical}. In the PI-ESN, the time-derivative of the network is approximated by a first-order forward Euler numerical scheme. In this work, we compute the derivative at machine precision through automatic differentiation \cite{tensorflow2015-whitepaper,baydin2017automatic}. 

In the API-ESN, the data for the observed state, $\textbf{x}$, updates the state of the high-dimensional reservoir, $\textbf{r}\in\mathbb{R}^{N_r}$, which acts as the memory of the network. At the $i$-th time step, $\textbf{r}(t_i)$ is a function of its previous value, $\textbf{r}(t_{i-1})$, and the current input, $\textbf{x}(t_i)$. The output is the predicted state at the next time step: $\mathbf{\hat{y}}(t_i)=[\mathbf{\hat{x}}(t_{i+1});\mathbf{\hat{h}}(t_{i+1})]\in\mathbb{R}^{N_{y}}$.
It is the linear combination of $\textbf{r}(t_i)$ and $\textbf{x}(t_i)$
\begin{equation}
\label{state_step}
        \textbf{r}(t_i) = \textrm{tanh}\left(\mathbf{W}_{\mathrm{in}}[\textbf{x}(t_i);b_{\mathrm{in}}]+
        \mathbf{W}\textbf{r}(t_{i-1})\right); \quad \mathbf{\hat{y}}(t_i) = \mathbf{W}_{\mathrm{out}}[\mathbf{r}(t_i);\mathbf{x}(t_i);1]
\end{equation}
where $\mathbf{W}\in\mathbb{R}^{N_r\times N_r}$ is the state matrix, $\mathbf{W}_{\mathrm{in}}\in\mathbb{R}^{N_r\times (N_x+1)}$ is the input matrix, $\mathbf{W}_{\mathrm{out}}\in\mathbb{R}^{ N_{y}\times(N_r+N_x+1)}$ is the output matrix and $b_{\mathrm{in}}$ is the input bias.
The input matrix, $\mathbf{W}_{\mathrm{in}}$, and state matrix, $\mathbf{W}$, are sparse, randomly generated and fixed. These are constructed in order for the network to satisfy the echo state property \cite{lukovsevivcius2012practical}. The input matrix, $\mathbf{W}_{\mathrm{in}}$, has only one element different from zero per row, which is sampled from a uniform distribution in $[-\sigma_{\mathrm{in}},\sigma_{\mathrm{in}}]$, where $\sigma_{\mathrm{in}}$ is the input scaling. The state matrix, $\textbf{W}$, is an Erdős-Renyi matrix with average connectivity $\langle d\rangle$. This means that each neuron (each row of $\mathbf{W}$) has on average only $\langle d\rangle$ connections (non-zero elements). The value of the non-zero elements is obtained by sampling from an uniform distribution in $[-1,1]$; the entire matrix is then scaled by a multiplication factor to set its spectral radius, $\rho$. The only trainable weights are those in the the output matrix, $\mathbf{W}_{\mathrm{out}}$. 
The first $N_x$ rows of the output matrix,  $\mathbf{W}_{\textrm{out}}^{(x)}$, are computed through Ridge regression on the available data for the observed state by solving the linear system
\begin{equation}
\label{ridge}
    \left(\mathbf{R}\mathbf{R}^T + \gamma \mathbf{I}\right)\mathbf{W}_{\mathrm{out}}^{(x)^{T}} 
    =\mathbf{R}\mathbf{X}^T,
\end{equation}
where $\mathbf{X}\in\mathbb{R}^{N_{x}\times N_{t}}$ and $\mathbf{R}\in\mathbb{R}^{(N_{r}+N_x+1)\times N_{t}}$ are the horizontal concatenation of the observed states, $\mathbf{x}$, and associated reservoir states, $[\mathbf{r};\mathbf{x};1]$, respectively; $\gamma$ is the Tikhonov regularization factor and $\mathbf{I}$ is the identity matrix \cite{lukovsevivcius2012practical}. The last $N_h$ rows of the output matrix, $\mathbf{W}_{\mathrm{out}}^{(h)}$, are initialized by solving (\ref{ridge}), where we embed prior knowledge of the physics by substituting $\mathbf{X}$ with $\mathbf{H}\in\mathbb{R}^{N_{h}\times N_{t}}$, whose rows are constants and equal to the components of the estimate of the mean of the hidden state, $\mathbf{\overline{h}}\in\mathbb{R}^{N_h}$. 
To train $\mathbf{W}_{\mathrm{out}}^{(h)}$ only, we minimize the loss function, $\mathcal{L}_{\mathrm{Phys}}$
\begin{equation}
\mathcal{L}_{\mathrm{Phys}} = \frac{1}{N_tN_y}\sum_{j=1}^{N_t}||\mathbf{\dot{\hat{y}}}(t_j)-\textbf{f}(\mathbf{\hat{y}}(t_j))||^2, \label{l_phys}
\end{equation}
where $||\cdot||$ is the $L_2$ norm; 
$\mathcal{L}_{\mathrm{Phys}}$ is the Mean Squared Error between the time-derivative of the output, $\mathbf{\dot{\hat{y}}}$, and the right-hand side of the governing equations evaluated at the output, $\mathbf{\hat{y}}$.
To compute $\mathbf{\dot{\hat{y}}}$, we need to differentiate $\mathbf{\hat{y}}$ with respect to $\mathbf{x}$, because the time dependence of the network is implicit in the input, $\mathbf{x}(t)$, i.e. $\frac{d\mathbf{\hat{y}}}{dt} = \frac{\partial\mathbf{\hat{y}}}{\partial \mathbf{x}}\frac{d\mathbf{x}}{dt}$. The fact that $\mathbf{\dot{\hat{y}}}$ is a function of $\mathbf{\dot{x}}$ means that the accuracy of the derivative of the output is limited by the accuracy of the derivative of the input. In this work, we compute $\mathbf{\dot{x}}$ exactly using the entire state to evaluate $\mathbf{f}(\mathbf{y})$ in (\ref{dyn_sys}). Furthermore, $\mathbf{\hat{y}}$ depends on all the inputs up to the current input due to the recurrent connections between the neurons. In Echo State Networks, we have the recursive dependence of the reservoir state, $\mathbf{r}$, with its previous values, i.e.,  omitting the input bias for brevity
\begin{equation}
\label{exp_state}
\SMALLL{
     \textbf{r}(t_i) = \textrm{tanh}{\bigg(}\mathbf{W}_{\mathrm{in}}\mathbf{x}(t_i)+\mathbf{W}\textrm{tanh}{\Big (}\mathbf{W}_{\mathrm{in}}\mathbf{x}(t_{i-1})+\mathbf{W}\textrm{tanh}(\mathbf{W}_{\mathrm{in}}\mathbf{x}(t_{i-2})+\dots){\Big)}{\bigg)}.
     }
\end{equation}
Because of this, the  time-derivative of the current output has to be computed with respect to all the previous inputs in the training set
\begin{equation}
\mathbf{\dot{\hat{y}}}(t_i) = \frac{d\mathbf{\hat{y}}}{dt}{\bigg |}_{t_i} = \frac{\partial\mathbf{\hat{y}}(t_i)}{\partial \textbf{x}(t_i)}\frac{d\textbf{x}}{dt}{\bigg |}_{t_i} + \frac{\partial\mathbf{\hat{y}}(t_i)}{\partial \textbf{x}(t_{i-1})}\frac{d\textbf{x}}{dt}{\bigg |}_{t_{i-1}} + \frac{\partial\mathbf{\hat{y}}(t_i)}{\partial \textbf{x}(t_{i-2})}\frac{d\textbf{x}}{dt}{\bigg |}_{t_{i-2}} + \ldots
\end{equation}
This is computationally cumbersome. To circumvent this extra computational cost, we compute $\mathbf{\dot{\hat{y}}}$ through the derivative of the reservoir's state, $\mathbf{\dot{r}}=\frac{\partial\mathbf{r}}{\partial \mathbf{x}}\frac{d\mathbf{x}}{dt}$. By differentiating (\ref{state_step}) with respect to time, we obtain 
$
  \mathbf{\dot{\hat{y}}} = \mathbf{W}_{\mathrm{out}}[\mathbf{\dot{r}};\mathbf{\dot{x}};0]
$. Because $\mathbf{\dot{r}}$ is independent of $\mathbf{W}_{\mathrm{out}}$, $\mathbf{\dot{r}}$ is fixed during training. Hence, the automatic differentiation of the network is performed only once during initialization. We compute $\mathbf{\dot{r}}$ as a function of the current input and previous state as the network evolves
\begin{equation}
\mathbf{\dot{r}}(t_i) = \frac{d\mathbf{r}}{dt}{\bigg |}_{t_i} =
\frac{\partial\mathbf{r}(t_i)}{\partial \mathbf{x}(t_i)}\frac{d\mathbf{x}}{dt}{\bigg |}_{t_i} + \frac{\partial\mathbf{r}(t_i)}{\partial \mathbf{r}(t_{i-1})}\frac{d\mathbf{r}}{dt}{\bigg |}_{t_{i-1}},
\end{equation}
in which we initialize $\mathbf{\dot{r}}(t_0)=0$ and $\mathbf{r}(t_0)=0$ at the beginning of the washout interval (the washout interval is the initial transient of the network, during which we feed the inputs without recording the outputs in order for the state of the network to be uniquely defined by the sequence of the inputs \cite{lukovsevivcius2012practical}). 



\section{Reconstruction of hidden states in a chaotic system}

\label{sec:Results}

We study the Lorenz system \cite{lorenz1963deterministic}, which is a prototypical chaotic system that models Rayleigh–Bénard convection
\begin{equation}
\label{lorenz}
    \dot{\phi}_1 = \sigma_L(\phi_2-\phi_1), \qquad
    \dot{\phi}_2 = \phi_1(\rho_L - \phi_3) - \phi_2, \qquad 
    \dot{\phi}_3 = \phi_1\phi_2 - \beta_L \phi_3,
\end{equation}
where the parameters are $[\sigma_L,\beta_L,\rho_L] = [10,8/3,28]$. To obtain the data, we integrate the equation through the implicit adaptive step scheme of the function {\tt odeint} in the {\tt scipy} library. The training set consists of $N_t=10000$ points with step $\Delta t=0.01$LTs, where a Lyapunov Time (LT) is the inverse of the leading Lyapunov exponent $\Lambda$ of the system, which, in turn, is the exponential rate at which arbitrarily close trajectories diverge. In the Lorenz system, $\Lambda = \textrm{LT}^{-1} \simeq 0.906$.
We use networks of variable sizes from $N_r=100$ to $N_r=1000$, with parameters $\langle d\rangle=20$, $\gamma=10^{-6}$ and $\rho=0.9$ \cite{lu2017reservoir}. We set $\sigma_{\mathrm{in}}=0.1$, $b_{\mathrm{in}}=10$, and $\overline{h}=10$, to take into account the order of magnitude of the inputs ($\sim10^1$). We train the networks using the Adam optimizer~\cite{kingma2014adam} with initial learning rate $l=0.1$, which we decrease during optimization to prevent the training from not converging to the optimal weights due to large steps of the gradient descent. 

\begin{figure}
    \centering
    \includegraphics[width=1.0\textwidth]{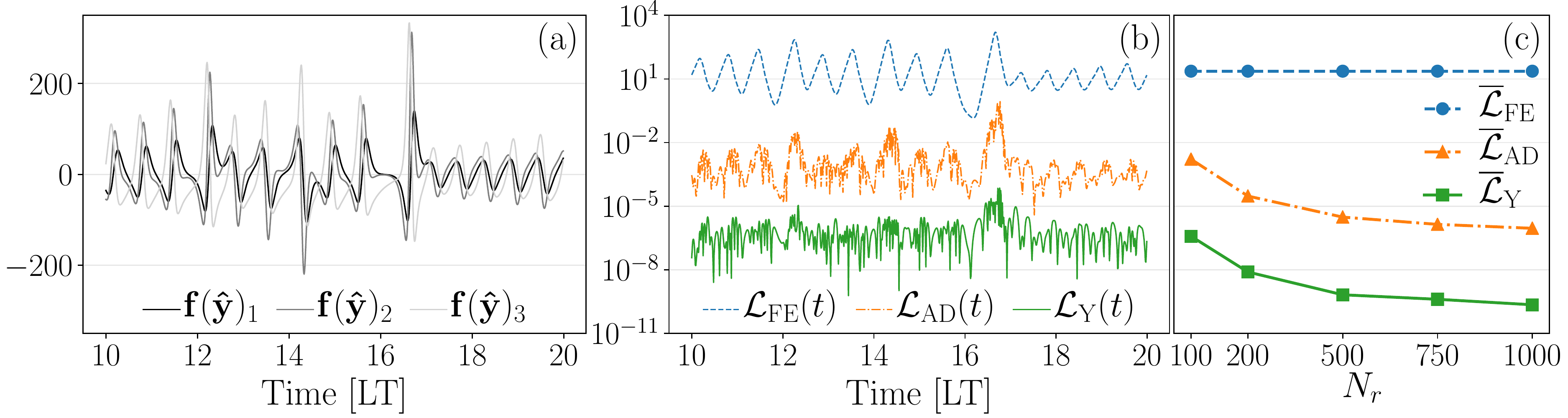}
    \caption{Time-derivative of the output for $N_r=100$ in an interval of the training set, (a). Squared norm of the error of the output derivative, $\mathcal{L}_{\mathrm{FE}}$ and $\mathcal{L}_{\mathrm{AD}}$ , and output, $\mathcal{L}_{\mathrm{Y}}$, in the same interval, (b). Mean Squared Error of the output derivative, $\overline{\mathcal{L}}_{\mathrm{FE}}$ and $\overline{\mathcal{L}}_{\mathrm{AD}}$ , and output, $\overline{\mathcal{L}}_{\mathrm{Y}}$, as a function of the reservoir size, (c).}
    \label{AD_acc}
\end{figure}

In Fig. \ref{AD_acc}, we compare the accuracy of the Automatic Differentiation (AD) derivative of the API-ESN with respect to the first-order Forward Euler (FE) approximation of the PI-ESN \cite{doan2020learning}. Here, we study the case where the entire state is known, $\mathbf{y}=\mathbf{x}$, to be able to compute the true derivative of the predicted state, $\mathbf{f}(\mathbf{\mathbf{\hat{y}}})$,  (\ref{dyn_sys}). In plot (a), we show $\mathbf{f}(\mathbf{\mathbf{\hat{y}}})$ in an interval of the training set for $N_r=100$. In plot (b), we show in the same interval the squared norm of the error for FE, $\mathcal{L}_{\mathrm{FE}}$, and AD, $\mathcal{L}_{\mathrm{AD}}$, with respect to $\mathbf{f}(\mathbf{\mathbf{\hat{y}}})$. In addition, we show the squared norm of the error, $\mathcal{L}_{\mathrm{Y}}$, of the output, $\mathbf{\hat{y}}$, with respect to the data, $\mathbf{y}$.  Because FE and AD share the same $\mathbf{\hat{y}}$, $\mathcal{L}_{\mathrm{Y}}$ is the same for the two networks. In plot (c), we show the time average of the squared norms, indicated by the overline, as a function of the size of the reservoir. In the case of $\overline{\mathcal{L}}_{\mathrm{FE}}$ and $\overline{\mathcal{L}}_{\mathrm{AD}}$, they coincide with $\mathcal{L}_{\mathrm{Phys}}$ (\ref{l_phys}).  
AD is four to seven orders of magnitude more accurate than FE. The error of the FE numerical approximation is dominant from $N_r=100$. This prevents its accuracy from increasing in larger networks.  

\begin{figure}
    \centering
    \includegraphics[width=1.\textwidth]{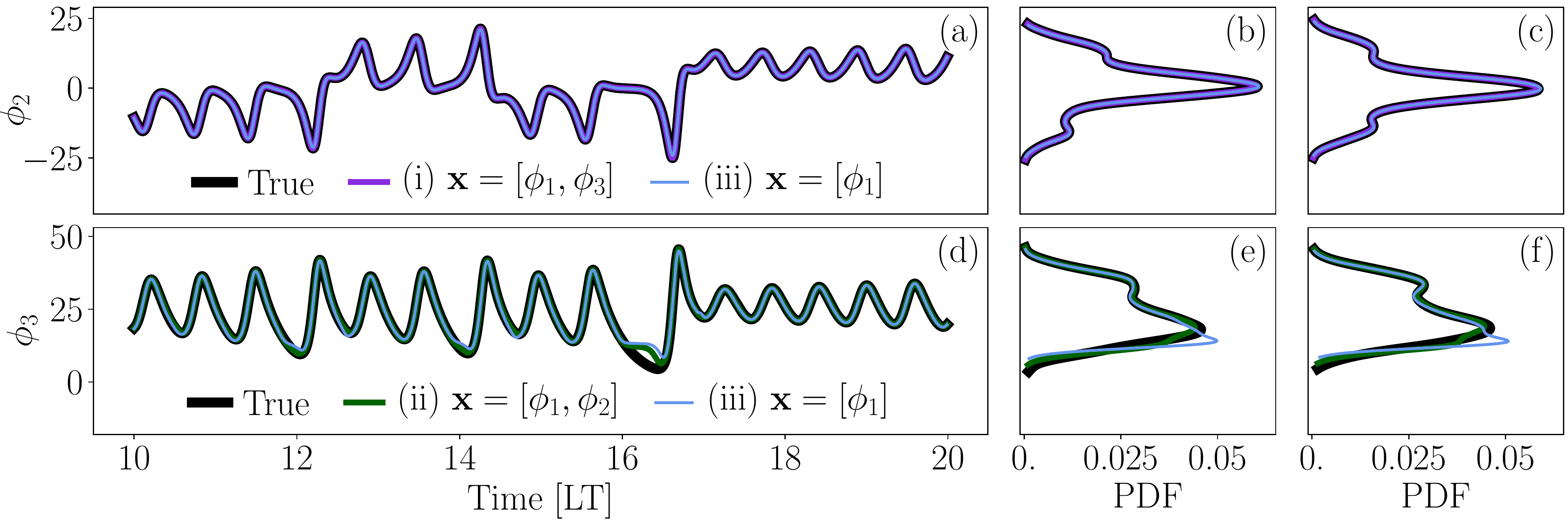}
    \caption{Reconstruction of the hidden variables for $N_r=1000$ in an interval of the training set, (a) and (d), Probability Density Function (PDF) in the training, (b,e), and test set, (c,f). Reconstruction of $\mathbf{h}=[\phi_2]$ (top row) and $\mathbf{h}=[\phi_3]$ (bottom row): testcase (i) is shown in (a-c), testcase (ii) is shown in (d-f), and testcase (iii) is shown in (a-f).}
    \label{Var_Rec}
\end{figure}

We use the API-ESN to reconstruct the hidden states in three testcases: (i) reconstruction of $\mathbf{h}=[\phi_2]$ given $\mathbf{x}=[\phi_1;\phi_3]$; (ii) reconstruction of $\mathbf{h}=[\phi_3]$ given $\mathbf{x}=[\phi_1;\phi_2]$; (iii) reconstruction of  $\mathbf{h}=[\phi_2;\phi_3]$ given $\mathbf{x}=[\phi_1]$.
We choose (i) and (ii) to highlight the difference in performance when we reconstruct different variables, and (iii) to compare the reconstruction of the states $\phi_2$ and $\phi_3$ when fewer observed states are available. 
We reconstruct the hidden states in the training set and in a $10000$ points test set subsequent to the training set. The reconstructed states in an interval of the training set for networks of size $N_r=1000$ are shown in plots (a,d) in Fig. \ref{Var_Rec}. The network is able to reconstruct satisfactorily the hidden states. The accuracy deteriorates only in the large amplitude oscillations of $\phi_3$, (d). To visualize the global performance over both the training set and the test set, we plot the Probability Density Functions (PDF) in (b,e), and (c,f), respectively. The PDFs are reconstructed with similar accuracy between the two sets. Interestingly, the increased difficulty in reconstructing $\phi_3$ is due to the dynamical the system's equations (rather than the network's ability to learn). Indeed, $\phi_3$ appears in only two of the three equations (\ref{lorenz}), whereas $\phi_2$ appears in all the equations and it is a linear function in the first equation. In other words, we can extract more information for $\phi_2$ than for $\phi_3$ from the constraint of the physical equations, which means that the network can better reconstruct the dynamics of $\phi_2$ vs. $\phi_3$. In the lower part of large amplitude oscillations of $\phi_3$ in particular, we have small values for the derivatives, so that the error in the governing equations is small.

\begin{figure}
    \centering
    \includegraphics[width=1.\textwidth]{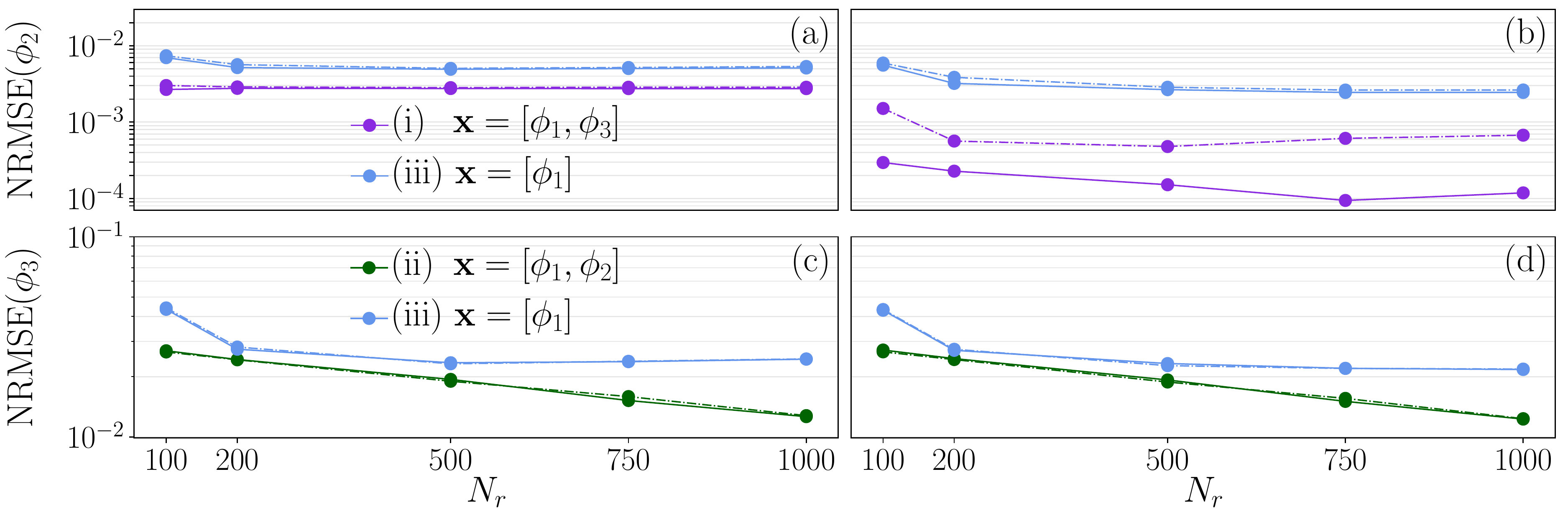}
    \caption{NRMSE for in the training (continuous lines) and test (dash-dotted lines) sets as a function of the size of the reservoir. The reconstruction is performed through forward Euler, (a,c), and automatic differentiation, (b,d). Reconstruction of $\mathbf{h}=[\phi_2]$, (top row) and $\mathbf{h}=[\phi_3]$ (bottom row): testcase (i) is shown in (a,b), testcase (ii) is shown in (c,d), and testcase (iii) is shown in (a-d).}
    \label{loss_bar}
\end{figure}

To quantitatively assess the reconstruction, we compute for each component, $h_i$, of the hidden state, $\mathbf{h}$, the Normalized Root Mean Squared Error: $\textrm{NRMSE}(h_i) = \sqrt{N^{-1}\sum_{j}^{N}(\hat{h}_i(t_j)-h_i(t_j))^2}/( \max(h_i)   -  \min(h_i))$;
where $( \allowbreak \max(h_i)  \allowbreak - \min(h_i))$ is the range of $h_i$. In Fig. \ref{loss_bar}, we show the values of the NRMSE for different sizes of the reservoir, $N_r$, in the training (continuous lines) and test (dash-dotted lines) sets for FE, (a,c), and AD, (b,d). The error of the FE approximation dominates the reconstruction of $\phi_2$, (a), while AD produces NRMSEs up to one order of magnitude smaller, (b). In the reconstruction of $\phi_3$, (c,d), AD and FE perform similarly because of the dominant error related to the mathematical structure of the dynamical system's equations, as argued for Fig.~\ref{Var_Rec} (e,f). In general, the values of the NRMSE are similar between the training and test sets, which indicates that the reconstruction also works  on unseen data. There is a difference between the two sets only in the reconstruction of $\phi_2$ given $\mathbf{x}=[\phi_1;\phi_3]$, (b). In this case, the NRMSE in the test set is smaller than the NRMSE we obtain when using data on the hidden state to train the network, through the network described in \cite{lu2017reservoir} (results not shown). In addition, we compare the reconstruction of the states when fewer observed states are available. When only $\phi_1$ is available ($\mathbf{x}=[\phi_1]$), the larger error on $\phi_3$ limits the accuracy on the reconstruction on $\phi_2$. This results in a reconstruction of $\phi_3$, (d), with similar accuracy to the case where $\mathbf{x}=[\phi_1;\phi_2]$, while there is a larger error in the reconstruction of $\phi_2$, (b), with respect to the case where $\mathbf{x}=[\phi_1;\phi_3]$.   


\section{Conclusions and future directions}
We propose the Automatic-differentiated Physics-Informed Echo State Network (API-ESN) to leverage the knowledge of the governing equations in an echo state network. 
We use automatic differentiation to compute the exact time-derivative of the output, which is shown to be a function of all the network's previous inputs
through the recursive time dependence intrinsic in the neurons' recurrent connections. Albeit this long time-dependence in the past would make the computation of the time-derivative computationally cumbersome, 
we eliminate this cost by computing the derivative of the reservoir's state. 
We apply the API-ESN to a prototypical chaotic system to reconstruct the hidden states in different datasets both in the training points and on unseen data. We compare the API-ESN to the forward Euler approximation of the Physics-Informed Echo State Network. We obtain a Normalized Mean Squared Error up to one order of magnitude smaller in the reconstruction of the hidden state. 
Future work will focus on using the API-ESN to reconstruct and predict (by letting the network evolve autonomously) hidden states from experimental data. 

\label{sec:conc}
%
%

%

\end{document}